\documentclass{ecai}
\usepackage{graphicx}
\usepackage{latexsym}

\usepackage{url}
\usepackage[hidelinks]{hyperref}
\usepackage[utf8]{inputenc}
\usepackage[small]{caption}
\usepackage{graphicx}
\usepackage{amsmath}
\usepackage{amsthm}
\usepackage{booktabs}
\usepackage{algorithm}
\usepackage{algorithmic}
\usepackage{array}
\usepackage{todonotes}
\usepackage{multirow}
\usepackage{amsthm,amsmath,amssymb} 
 
\usepackage{subcaption}

\newtheorem{proposition}{Proposition}

\theoremstyle{definition}
\newtheorem{definition}{Definition}

\newcommand{\mlp}{\ensuremath{\mathcal{M}}}
\newcommand{\weights}{\ensuremath{\mathcal{W}}}
\newcommand{\biases}{\ensuremath{\mathcal{B}}}
\newcommand{\outputf}{\ensuremath{\mathcal{O}}}

\newcommand{\domain}{\ensuremath{\mathcal{D}}}
\newcommand{\arguments}{\ensuremath{\mathcal{A}}}
\newcommand{\attacker}{\ensuremath{\mathrm{Att}}}
\newcommand{\supporter}{\ensuremath{\mathrm{Sup}}}
\newcommand{\baseScore}{\ensuremath{\beta}}
\newcommand{\strength}{\ensuremath{\sigma}}

\newcommand{\partition}{\ensuremath{\mathcal{P}}}
\newcommand{\dist}{\ensuremath{\delta}}

\newcommand{\agg}{\ensuremath{\operatorname{Agg}}}

\newcommand{\pointspace}{\ensuremath{\mathcal{S}}}
\newcommand{\dataset}{\ensuremath{
\Delta}}

\newtheorem{example}{Example}

\begin{document}

\begin{frontmatter}

\title{[Technical Report]\\SpArX: Sparse Argumentative Explanations \\ for Neural Networks}


\author[A]{\fnms{Hamed}~\snm{Ayoobi}\orcid{0000-0002-5418-6352}\thanks{Corresponding Author. Email: h.ayoobi@imperial.ac.uk}}
\author[B,A]{\fnms{Nico}~\snm{Potyka}}
\author[A]{\fnms{Francesca}~\snm{Toni}}

\address[A]{Department of Computing, Imperial College London, United Kingdom}
\address[B]{Cardiff University, United Kingdom}

\begin{abstract}
Neural networks (NNs) have various applications
in AI, but explaining
their decisions remains challenging. 
Existing approaches often focus on explaining how changing individual inputs affects NNs'
outputs. 
However, an explanation that is consistent
with the input-output behaviour of an NN is not 
necessarily faithful to the actual mechanics
thereof. In this paper, we exploit
relationships between \emph{multi-layer perceptrons} (MLPs)
and \emph{quantitative argumentation frameworks} (QAFs)
to create argumentative explanations for the mechanics of MLPs. 
Our \emph{SpArX} method first sparsifies the
MLP while maintaining as much of the original
structure as possible.
It then translates
the sparse MLP into an equivalent QAF to shed light on the underlying decision process of the MLP, producing  
 {\em global} and/or {\em local explanations}. We demonstrate experimentally
that SpArX can 
give more faithful explanations than existing approaches,
while simultaneously providing deeper insights
into the actual  
reasoning process of MLPs.
\end{abstract}

\end{frontmatter}

\section{Introduction}

The increasing use of black-box models like
neural networks (NNs) in autonomous intelligent systems raises concerns about their fairness, reliability and safety. To address
these concerns, the 
literature
puts forward various explainable AI approaches to 
render
NNs more transparent
, including model-agnostic approaches~\cite{LIME_2016should,lundberg2017unified},
and approaches tailored to the structure of NNs~\cite{
LRP,ZintgrafCAW17}.
However, they fail to capture the actual mechanics of the NNs 
and thus 
it 
is hard to evaluate how \emph{faithful} 
these approaches are to the NNs
~\cite{heo2019fooling,sanchez2020evaluating,rao2022towards}.

Some works advocate the use of formal, interpretable approaches for explainability~\cite{FormalXAI-22}. 
Specifically for NNs, recent work~\cite{wu2021optimizing} proposes
regularizing the training procedure of NNs so that they can be well
approximated by interpretable decision trees.
While this is an interesting direction,
evaluating the faithfulness of the decision trees to the NN
remains a challenge. 
Other recent work unearths formal relationships between 
NNs in the form of 
multi-layer perceptrons (MLPs) 
and symbolic reasoning 
with 
\emph{quantitative argumentation frameworks (QAFs)} 
~\cite{Potyka_21,Ay:ABL,Ay:AABL, Ay:ExplainWhatYouSee, Ay:CASE, Ay:Local-HDP} or weighted conditional
knowledge bases ~\cite{Giordano2021}.
The formal relationships 
indicate that these approaches may pave the way towards potentially 
more faithful explanations 
than approximate abstractions such as decision trees. 

In this paper,  we provide explanations for MLPs leveraging on their formal relationships with QAFs in \cite{Potyka_21}.
Intuitively, QAFs
represent arguments and
relations of attack or support between them as a 
graph, where nodes
represent arguments and edges 
relations
.
Various QAF formalisms have been studied 
over the
years, e.g. by~\cite{bipolar05,amgoud2008bipolarity,baroni2015automatic,rago2016discontinuity,amgoud2017acceptability,potyka2018Kr,mossakowski2018modular,potyka_modular_2019}.
As it turns out, every MLP
corresponds to a  QAF of a particular form 
under a particular
semantics 
~\cite{Potyka_21}.
This formal relationship between MLPs and QAFs
suggests that QAFs are
well suited to create faithful explanations for MLPs.
%
However, 
just reinterpreting an MLP as a
QAF 
would not give us a comprehensible explanation
because the QAF has the same size and density as the
original MLP.
In order to create  
faithful and comprehensible argumentative
explanations, we propose a two-step method
. We first \emph{sparsify} the MLP,
while maintaining as much of its mechanics as 
possible.
Then, we 
translate the sparse MLP into a QAF.
We call our method \emph{SpArX} (\emph{Sparse Argumentative eXplanations} for MLPs).
In principle, any existing 
compression method for NNs can be used for sparsification (e.g.~\cite{Yu_2022_WACV}). 
However, existing methods are not designed for
maintaining the mechanics of NNs towards explainability. 
We thus make the following contributions:
\begin{itemize}
    \item 
We propose a novel \emph{clustering method} 
for summarizing neurons based on their output-similarity.
The clustered neurons' parameters result from aggregating the
original parameters 
so
that 
their output 
is similar to the outputs of neurons that 
they 
summarize. 

\item We propose two families of
\emph{aggregation 
functions} 
 for aggregating the neurons in a cluster: the first gives
\emph{global explanations} (explaining the 
 MLP for all inputs) 
and the second gives
\emph{local explanations} (explaining the MLP for a target input). 
\item 
We 
conduct 
several experiments demonstrating
the viability of our 
SpArX method for MLPs and its competitiveness with respect to other methods in terms of (i) conventional notions of \emph{input-output faithfulness}
of explanations and  (ii) novel notions of \emph{structural faithfulness},  while (iii) shedding some light on the tradeoff between
faithfulness and comprehensibility understood in terms of a notion of \emph{cognitive complexity}
, important towards human usability of explanations with SpArX. The code is publicly available\footnote{\url{https://github.com/H-Ayoobi/SpArX}}. 
\end{itemize}
Overall, we show that formal relationships between black-box machine learning 
(
with NNs) and interpretable symbolic reasoning 
in 
QAFs can provide faithful and comprehensible explanations. 

This is an extended version of \cite{AyoobiECAI2023}, including Supplementary Material (SM) with proofs and additional details. 

\section{Related Work}

While MLPs are most commonly used in their
fully connected form, there has been 
increasing interest in learning
sparse NNs in recent years.
However, the focus is usually not on 
finding an easily interpretable network
structure, but rather on decreasing the 
risk for overfitting, memory and 
runtime complexity and the associated power consumption.
Existing approaches include regularization to encourage neurons with weight $0$ 
to be deleted ~\cite{LouizosWK18},
 pruning of edges ~\cite{Yu_2022_WACV}, compression ~\cite{Louizos_2017} 
and low rank approximation ~\cite{Tai_2016}. 
Interval NNs ~\cite{prabhakar2019abstraction}
summarize neurons in clusters based on
their parameters and 
consider interval outputs for the clustered
neurons to give lower and upper bounds
on the outputs.
We also summarize neurons in clusters,
but cluster neurons based on their output
and return an aggregated output instead
of an interval for cluster neurons.


Several approaches exist for obtaining argumentative explanations for a variety of models~\cite{argXAIsurvey}.
Some approaches use argumentation to explain models directly \cite{Ay:ABL}, others use argumentative counterparts of the models.
Some of them focus on NN explanations~\cite{DAX,LRPArgExpl21}, but they
are based on approximations of NNs (e.g. using Layerwise Relevance Propagation~\cite{LRP}), rather than summarization as in our method, and their faithfulness is difficult to ascertain.

Several existing methods make use of symbolic reasoning 
for providing explanations \cite{FormalXAI-22}. The explanations resulting from these methods (e.g. abduction-based explanations \cite{AbdX}, prime implicants \cite{PrimeImpl}, sufficient reasons \cite{SuffReas}, and majority reasons~\cite{MajorReas22})  faithfully capture the input-output behaviour of the explained models rather than their mechanics.  Other methods extract logical rules as explanations for machine learning models, including NNs \cite{LORE19,LeiteKR22}, but again  focus on explanations that are only input-output faithful.


\section{Preliminaries}
\label{sec:pre}
Intuitively, a multi-layer perceptron (MLP) is a layered acyclic graph that processes its input by propagating
it through the layers.
Formally, we describe MLPs as follows.

\begin{definition}[Multi-Layer Perceptron (MLP)]
An \emph{MLP} $\mlp$ is a tuple $
(V, E, \biases, \weights, \varphi)$. 
$(V,E)$ is a directed graph.
$V \!=\! \uplus_{l=0}^{d+1} V_l$ consists of (ordered) layers of neurons; for $0\!\leq\! l\!\leq\! d+1$, $V_l = \{v_{l,i} \mid 1 \leq i \leq |V_l|\}$:
	 we call $V_0$ the \emph{input layer}, $V_{d+1}$ the \emph{output layer} and  $V_l$, for $1\leq l \leq d$, the $l$-th \emph{hidden layer};  $d$ is the \emph{depth} of the MLP.
$E \subseteq \bigcup_{l=0}^d \big(V_l \times V_{l+1} \big)$ is a set of edges between
 adjacent layers;
if $E = \bigcup_{l=0}^d \big(V_l \times V_{l+1} \big)$, then the MLP is called \emph{fully connected}.
$\biases = \{b^1, \dots, b^{d+1}\}$ is a set of  \emph{bias} vectors, where, for  $1\leq l \leq d+1$, $b^l \in \mathbb{R}^{|V_l|}$.
$\weights \!=\! \{W^0, \!\dots, \!W^d\}$ is a set of \emph{weight} matrices, where, for $1\!\leq \!l \!\leq \! d$, $W^l \!\in \! \mathbb{R}^{|V_{l+1}|\times|V_{l}|}$
 such that $W^l_{i,j}\!=\!0$ when 
 $(v_{l,j},v_{l+1,i}) \!\not\in \!E$.
$\varphi: \mathbb{R} \rightarrow \mathbb{R}$ is an
\emph{activation function}.
\end{definition}

An example of MLP is given later in Fig.~\ref{fig_XOR_example_summarized}a.
In order to process an \emph{input} $x \!\in\! \mathbb{R}^{|V_0|}$, the input layer of 
$\mlp$ is initialized with $x$. 
The 
input is then propagated forward through 
$\mlp$
to generate values at each subsequent layer and 
finally
an \emph{output} in the output layer. 
Formally, if the values at layer $l$ are $x_l \!\in \!\mathbb{R}^{|V_l|}$, then the values $x_{l+1} \!\in\! \mathbb{R}^{|V_{l+1}|}$ at the next layer are given
by $x_{l+1} \!=\! \varphi(W^l \, x_l + b^l)$, 
with
the activation function $\varphi$ 
applied component-wise.
We let 
$\outputf^\mlp_{x}\!: \!V \!\rightarrow \!\mathbb{R}$
denote the \emph{output function} of $\mlp$, assigning
to each neuron its 
value when the input $x$ is given.
That is, for $v_{0,i} \in V_0$, we let
$\outputf^\mlp_{x}(v_{0,i}) \!=\! x_i$ and, for $l\!>\!0$,
we let the \emph{activation value} of neuron 
$v_{l,i}$ be  $\outputf^\mlp_{x}(v_{l,i}) \!=\! 
\varphi(W^l \, \outputf^\mlp_{x}(V_{l-1}) + b^l)_i$, 
where $\outputf^\mlp_{x}(V_{l-1})$ denotes the vector 
obtained
from $V_{l-1}$ by applying $\outputf^\mlp_{x}$
component-wise. 
%

Every MLP can be seen as a
quantitative argumentation framework (QAF) ~\cite{Potyka_21}.
Intuitively, QAFs are \emph{edge-weighted} directed graphs,
where nodes represent \emph{arguments} and, similarly to ~\cite{mossakowski2018modular},  edges with negative weight represent
\emph{attack} and edges with positive weight represent \emph{support} relations between arguments.
Each argument is initialized with a \emph{base score} that
assigns an apriori \emph{strength} to the argument.
The strength of 
arguments is then updated iteratively
based on the strength values of attackers and supporters
until the values converge. 
In acyclic graphs corresponding to MLPs, this iterative process is equivalent to the forward propagation process in the MLPs ~\cite{Potyka_21}. 
Conceptually, strength values 
are from some 
\emph{domain} $\domain$ ~\cite{baroni2018many}.
As we focus on (real-valued) MLPs, we will assume $\domain \subseteq \mathbb{R}$. The exact domain depends on 
the activation function
, e.g. the logistic function
results in $\domain = [0,1]$, the hyperbolic tangent in $\domain = [-1,1]$ and ReLU in $\domain = [0,\infty]$.  
Formally, we describe QAFs as follows.
\begin{definition}[Quantitative Argumentation Framework (QAF)]
A \emph{QAF with domain} $\domain \subseteq \mathbb{R}$ is a tuple
$(\arguments, E, \baseScore, w)$ that consists of
\begin{itemize}
    \item 
    sets $\arguments$ of \emph{arguments}  and 
    $E \!\subseteq\! \arguments \!\times\! \arguments$ of \emph{edges} 
between 
arguments;
    \item a function $\baseScore\!:\! \arguments \! \rightarrow \! \domain$
assigning \emph{base scores} 
in
$\domain$
to 
all arguments;
    \item a function $w: E \rightarrow \mathbb{R}$
     assigning \emph{weights} in $\mathbb{R}$ to 
     all edges.
\end{itemize}
Edges with negative/positive weights are called \emph{attack}/\emph{support} edges
, denoted by $\attacker$/$\supporter$, respectively.
\end{definition}
The strength values of arguments are usually computed iteratively
using a two-step update procedure ~\cite{mossakowski2018modular}: first, an \emph{aggregation function} $\alpha$ 
aggregates the strength values of attackers and supporters;
then, an \emph{influence function} $\iota$ adapts the
base score. 
Examples of aggregation functions include product ~\cite{baroni2015automatic,rago2016discontinuity},
addition ~\cite{amgoud2017evaluation,potyka2018Kr} and maximum ~\cite{mossakowski2018modular},  with the influence function
defined accordingly to guarantee that strength values fall
in 
$\domain$.
%
Here, we focus on  
the aggregation and influence functions from 
\cite{Potyka_21}, to obtain QAFs simulating MLPs 
with a logistic activation function ~\cite{Potyka_21}.
The strength values of arguments are computed by the following iterative
procedure:
for every 
$a \in \arguments$, we let $s_a^{(0)} := \baseScore(a)$
be the initial strength value; the strength values are then updated by the 
next two steps repeatedly 
(where the auxiliary 
$\alpha_a^i$ carries the aggregate at iteration $i\geq 0$):
\begin{description}
\item[Aggregation:] 
$\alpha_a^{(i+1)} := \sum_{(b,a) \in E} w((b,a)) \cdot s_b^{(i)}$.
\item[Influence:] 
$s_a^{(i+1)} := \varphi_l\big(\ln(\frac{\baseScore(a)}{1- \baseScore(a)}) + \alpha_a^{(i+1)} \big)$,
where $\varphi_l(z) = \frac{1}{1 + \exp(-z)}$ is the logistic function.
\end{description}
The \emph{final strength} of argument $a$ is defined via the limit of $s_a^{(i)}$, for $i$ towards infinity.
Notably, the semantics given by this notion of final strength satisfies almost all desiderata for QAF semantics~\cite{Potyka_21}.

\section{From General MLPs to QAFs}
\label{sec:preplus}

Here we generalize
the connection between MLPs and QAFs beyond MLPs with logistic activation functions,
as follows. Assume that $\varphi: \mathbb{R} \rightarrow \domain$
is an activation function that is strictly monotonically
increasing. Examples include logistic, hyperbolic tangent 
and parametric ReLU activation functions.
Then $\varphi$ is invertible and $\varphi^{-1}: \domain \rightarrow \mathbb{R}$ is defined. 
We can then define the update function for an MLP with such activation function $\varphi$ by using the same aggregation function as before and using the
following influence function:
\begin{description}
\item[Influence:] 
$s_a^{(i+1)} := \varphi\big(\varphi^{-1}(\baseScore(a)) + \alpha_a^{(i+1)} \big)$.
\end{description}
Note that the previous definition of influence in Section~\ref{sec:pre}, from ~\cite{Potyka_21},
is a special case because $\ln(\frac{1}{1-x})$ is the
inverse function of the logistic function $\varphi_l(x)$.
Note also that the popular ReLU activation function $\varphi_{ReLU}(x) = \max(0, x)$ is not invertible because all non-positive numbers are mapped to $0$. However, for our purpose of translating MLPs to QAFs, we can define $$\varphi_{ReLU}^{-1}(x) = 
\begin{cases} 			x, & \text{if $x>0$; }\\            
0, & \text{otherwise}.             
\end{cases}$$

In order to translate an MLP $\mlp$ with activation function
$\varphi$ and input $x$ into a QAF $Q_{\mlp, x}$, 
we interpret every neuron $v_{l,i}$ as an abstract 
argument $A_{l,i}$. 
Edges in $\mlp$ with positive/negative weights are interpreted
as supports/attacks, respectively, in $Q_{\mlp, x}$. 
The base score of an argument $A_{0,i}$ associated with input
neuron $v_{0,i}$ is just the corresponding input value $x_i$.
The base score of the remaining arguments $A_{l,i}$ is $\varphi(b^l_i)$,
where $b^l_i$ is the bias of the associated neuron $v_{l,i}$.
\begin{proposition}
\label{prop:translation}
Let $\mlp$ be an MLP with an invertible activation function $\varphi$ or ReLU. Then,  for  every input $x$, 
the 
QAF $Q_{\mlp,x}$ satisfies 
$\outputf^\mlp_{x}(v_{l,i}) = \strength(A_{l,i})$, where
$\strength(A_{l,i})$ denotes the final
strength of $A_{l,i}$ in $Q_{\mlp,x}$.
{\rm (See the Supplementary Material (SM) in \cite{ayoobi2023} for a proof). }
\end{proposition}


\section{SpArX: Explaining MLPs with QAFs}
\label{sec:sparx}

Just translating an MLP into a
QAF may 
not give a comprehensible explanation
because the QAF has the same size and density as the
original MLP.
Thus, we first sparsify the MLP and then translate it into a QAF.
The sparsification should maintain as much of
the original 
MLP as possible to give 
faithful explanations.
To achieve this
, we exploit redundancies
in the MLP by replacing
neurons 
giving similar outputs with a single neuron that summarizes their joint effect.

Summarizing neurons in this way is a clustering
problem. 
Formally, a clustering problem is defined by a set of 
inputs from an abstract space $\pointspace$ and a distance
measure 
$\dist: \pointspace \times \pointspace \rightarrow \mathbb{R}_{\geq 0}$. The goal is to partition $\pointspace$
into clusters $C_1, \dots, C_K$ (where $\pointspace=\uplus_{i=1}^{K} C_i$)
such that the distance
between points within a cluster is 'small' and the distance
between points in different clusters is 'large'.
Finding an optimal clustering is NP-complete in many cases ~\cite{gonzalez1982computational}. 
Thus, we cannot expect to 
find an efficient algorithm that computes an optimal clustering
, but we can apply standard algorithms e.g. K-means \cite{kmeans}
to find a good
(but not necessarily optimal) clustering efficiently.

\begin{figure}[t]
\setlength\tabcolsep{1.5pt} 
\begin{tabular}{ m{3.1cm} c c} 
\vspace{-29mm}
\begin{tabular}{ c }
\includegraphics[width=0.99
 \linewidth]{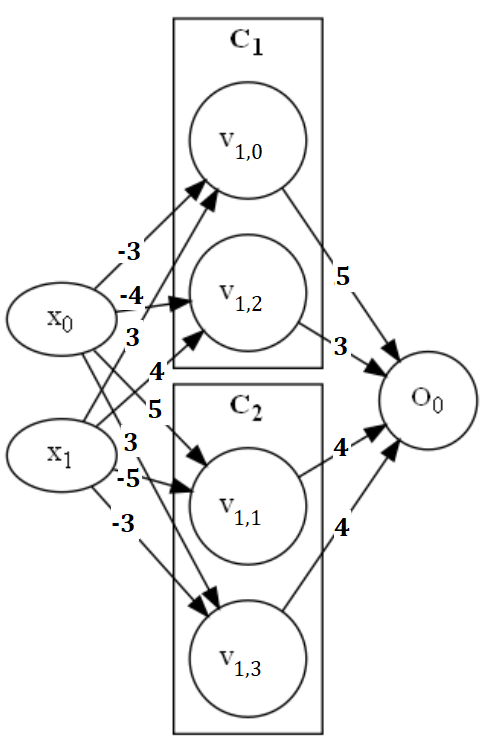}\\ (a)
 \end{tabular}
 & 
 \begin{tabular}{ c }
  \includegraphics[width=0.299\linewidth]{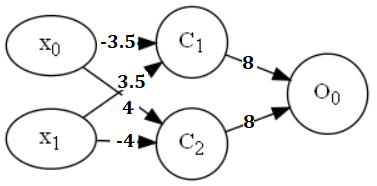} \\ (b)  \\
  \includegraphics[width=0.299\linewidth]{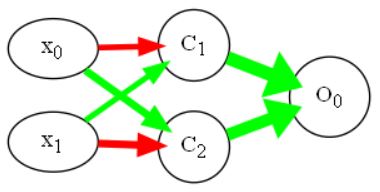} \\ (c) \\
  \includegraphics[width=0.2\linewidth]{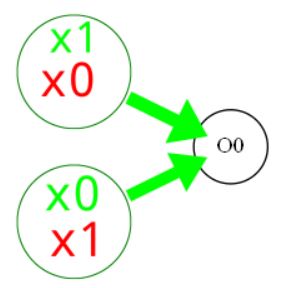} \\ (d)
  \end{tabular} 
  &
   \begin{tabular}{ c }
  \includegraphics[width=0.299\linewidth]{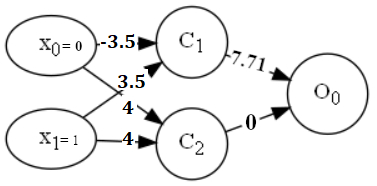} \\ (e)  \\
  \includegraphics[width=0.299\linewidth]{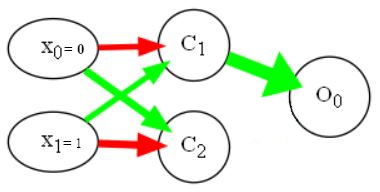} \\ (f) \\
  \includegraphics[width=0.2\linewidth]{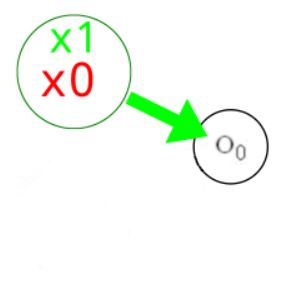} \\ (g)
  \end{tabular}
\end{tabular} 

\caption{a) MLP for XOR, with cluster-neurons $C_1,C_2$. b) 
Clustered MLP for  global explanation
. c) Global explanation 
as a QAF. d) Word-cloud representation 
for the global explanation
. e) Clustered MLP for the local explanation for $x_0\!\!=\!\!0,x_1\!\!=\!\!1$. f) Local explanation as a QAF. g) Word-cloud representation for the local explanation. }
\label{fig_XOR_example_summarized}
\end{figure}

In our setting, 
$\pointspace$ is the set $V_l$ of neurons in 
layer $0 \!< \!l\!< \!d\!+\!1$ and the distance between 
neurons can be
defined as the difference between their outputs
for  inputs in a given dataset \dataset\ (e.g. the training dataset):
\begin{equation}
    \dist(v_{l,i}, v_{l,j}) = \sqrt{\sum_{x \in \dataset}(\outputf^\mlp_{x}(v_{l,i}) - \outputf^\mlp_{x}(v_{l,j}))^2}. 
    \label{eq:dist}
\end{equation}
After clustering, 
we have a partitioning 
$\partition = \uplus_{l=1}^d \partition_l$
of (the hidden layers of) our MLP $\mlp$, where $\partition_l = \{C^l_1, \dots, C^l_{K_l}\}$ is the clustering of the $l$-th layer, that is,
$V_l = \uplus_{i=1}^{K_l} C^l_i$.
We use the clustering to create a corresponding 
\emph{clustered MLP} $\mu$ whose neurons correspond
to clusters in the original MLP $\mlp$.
We call these neurons $\emph{cluster-neurons}$
and denote them by $v_C$, where $C$ is the associated cluster. 
Then:
\begin{definition}[Graphical Structure of Clustered MLP]\label{def:global_explanation}
Given MLP $\mlp
$ and clustering
$\partition \!= \!\uplus_{l=1}^d \partition_l$ of $\mlp$, 
the \emph{graphical structure of the corresponding clustered MLP} $\mu$
is a directed graph
$(V^\mu,E^\mu)$ 
with
	 \begin{itemize}
  \item $V^\mu \!=\! \uplus_{l=0}^{d+1} V^\mu_l$ consists of (ordered) layers of cluster-neurons 
such that:
 \begin{enumerate}
    \item the input layer $V^\mu_0$ 
    consists of a singleton cluster-neuron 
    $v_{\{v_{0,i}\}}$ for every input neuron $v_{0,i} \in V_0$;
    \item 
    the 
    $l$-th hidden layer of $\mu$ (for $0 \!< \!l \!< \! d+1$) consists of one cluster-neuron
    $v_C$
    for every cluster $C \in \partition_l$;
    \item the output layer $V^\mu_{d+1}$ 
    consists of a singleton cluster-neuron 
    $v_{\{v_{d+1,i}\}}$ for every output neuron $v_{d+1,i} \!\in \!V_{d+1}$;
    \end{enumerate}
\item $E^\mu = \bigcup_{l=0}^d \big(V^\mu_l \times V^\mu_{l+1} \big)$
.
\end{itemize}
\end{definition}
\begin{example}
\label{ex:XOR}
Consider the MLP in Fig.~\ref{fig_XOR_example_summarized}.a, 
trained to approximate the
XOR function 
from the dataset $\dataset\!=\!\{
(0, 0),(0, 1), (1, 0), (1, 1)\}
$ with target outputs
$
0, 1, 1, 0
$, respectively. 
The  activation values of the hidden neurons for the four 
inputs are $
(0, 0, 0, 0),$ $(1.7, 0, 1.8, 0),(0, 2.3, 0, 1.5),(0, 0, 0, 0)
$, respectively.  
Applying $K$-means 
with $\dist$ as in Eq.~\ref{eq:dist} and $K\!=\!2$ for the hidden layer
results in 
clusters $C_1, C_2$ (indicated by rectangles in the figure). Fig.~\ref{fig_XOR_example_summarized}.b shows the graphical structure of the corresponding clustered MLP. 
\end{example}
We define \emph{global explanations} (for all inputs) 
and \emph{local explanations} 
(for specific inputs) 
for MLPs  by translating  their corresponding clustered MLPs into QAFs, leveraging on the formal correspondence in Prop.~\ref{prop:translation}. By doing so, 
we see QAFs themselves, equipped with the `final strength' semantics from Section~\ref{sec:preplus}, as explanations. 
Using the terminology of~\cite{argXAIsurvey}, thus, our approach is in the spirit of \emph{post-hoc approximate} approaches for argumentative explainable AI.
Below and in Section~\ref{sec:cognitive} we will discuss how QAFs can be tailored to human users to support varied explanatory experiences.  

The clustered MLPs for global and local explanations share the same 
 graphical structure but 
differ in  the parameters
of the cluster-neurons, that is, 
(i) the biases of cluster-neurons and 
(ii) the weights for edges between cluster-neurons.
We define these parameters in terms of \emph{aggregation functions}, 
 specifically a \emph{bias aggregation function} $\agg^b: \partition \rightarrow \mathbb{R}$, 
mapping  clusters to biases, and
an \emph{edge aggregation function}
$\agg^e: \partition\times\partition  \rightarrow \mathbb{R}\cup \{\bot\}$, mapping pairs of clusters to weights if the pairs correspond to edges in $\mu$, or $\bot$ otherwise. 
Given any concrete such aggregation functions (as defined later),  
the parameters of $\mu$ can be defined as
follows.

\begin{definition}[Parameters of Clustered MLP]
Given an MLP $\mlp
$, let $(V^\mu,E^\mu)$ be the graphical structure of the corresponding clustered MLP $\mu$. Then, for 
bias and edge aggregation functions $\agg^b$ and  
$\agg^e$, 
respectively, 
$\mu$ is $(V^\mu,E^\mu,\biases^\mu,\weights^\mu,\varphi)$
with \emph{parameters} $\biases^\mu,\weights^\mu$
as follows:
\begin{itemize}
    \item for every cluster-neuron $v_C \in V^\mu$,
    the bias (in $\biases^\mu$) of $v_C$  is
     $\agg^b(C)$;
    \item for every 
    edge $(v_{C_1}, v_{C_2})\in E^\mu$, the weight (in $\weights^\mu$) of the edge  is   
 $\agg^e((C_1, C_2))$.
\end{itemize}
\end{definition}

\subsection{Sparse Argumentative Global Explanations}

We 
use the following aggregation functions, which minimize
the deviation (with respect to the least-squares error) of bias and weights
of  cluster-neurons and the neurons
they contain (as we explain in the 
SM in \cite{ayoobi2023}).
\begin{definition}[Global Aggregation Functions]\label{def:aggregations_global}
The \emph{average bias and edge aggregation functions} are,  respectively:
\begin{eqnarray*}
    && \agg^b(C) = \frac{1}{|C|} \sum_{v_{l,i} \in C} b^l_i; 
    \\
    && \agg^e((C_1, C_2)) = 
    \sum_{v_{l,i} \in C_1} 
    \frac{1}{|C_2|} \sum_{v_{l+1,j} \in C_2} W^l_{j,i}.
\end{eqnarray*}
\end{definition}
The 
former
simply averages
the biases of neurons in the cluster.
For the latter, intuitively, the weight of an edge between cluster-neurons
$v_{C_1}$ and $v_{C_2}$ has to capture the effects of 
all neurons summarized in $C_1$ on neurons summarized in $C_2$.
Every neuron in $C1$ is connected to all neurons in $C2$, thus the aggregated weight between them encapsulates all the weights between $C1$ and $C2$.
As $v_{C_2}$ acts as a replacement of all neurons in 
$C_2$, it has to aggregate their activation. 
We achieve this aggregation by averaging again.

The following example illustrates global explanations drawn from clustered MLPs via their understanding as QAFs.
\begin{example}
\label{ex:globalclouds}
The QAF corresponding 
 to the clustered MLP  
 from Fig.~\ref{fig_XOR_example_summarized}.b can be visualised as in Fig.~\ref{fig_XOR_example_summarized}.c, where attacks are 
 in red, supports in green, and the thickness of the edges reflects their weight. 
 The same QAF can be visualized in different ways, e.g. to emphasize the role of each cluster-neuron. 
For instance, we can use a word-cloud representation as in Fig.~\ref{fig_XOR_example_summarized}.d (showing, e.g., that $x_0$ and $x_1$ play a negative  and positive role, respectively, towards $C_1$, which supports the output). This word-cloud representation gives full insights into the reasoning of the MLP (with  the 
learned  rule ($\overline{x_0} \wedge x_1) \vee (x_0 \wedge \overline{x_1}$)).
\end{example}

In general, word-cloud representations of cluster-neurons can be systematically generated 
by associating each cluster with a set of ``most relevant'' features, as follows.
Every input neuron is associated with the corresponding
feature; cluster-neurons in the first hidden layer are associated with the set of ``most relevant'' features from the input layer; cluster-neurons in the second layer are associated with the ``most relevant'' sets of sets of features that correspond to the most ``relevant features'' from the previous layer; and so on. 
We can measure ``relevance'' by the magnitude of its influence, which is the absolute value of edge weights for global explanations. 
For each word-cloud, 
the $k$  most relevant features  can be selected. As in Fig.~\ref{fig_XOR_example_summarized}.d, these features can be shown in green (red) if their influence is positive (negative, respectively). Also, the font size of features in word-clouds can be  proportional to
the magnitude of their influence.

\subsection{Sparse Argumentative Local Explanations}
While global explanations attempt to
faithfully explain the behaviour of the MLP
on all inputs
, our local explanations focus
on the behaviour in the neighborhood of 
the input $x $ from the dataset $\dataset$, similarly to LIME ~\cite{LIME_2016should}.
To do so, we generate random neighbors of 
$x$ to obtain a \emph{sample dataset} $\dataset'$, 
and weigh them 
with an exponential kernel from LIME~\cite{LIME_2016should}, 
assigning lower weight to a sample $x' \in \dataset'$ that is
further away from the target $x$:

\hspace*{2cm}  \(   \pi_{x', x} = exp(-D(x', x)^2/\sigma^2)
\)
\\
with $D$ the 
Euclidean distance
, $\sigma$ the width of the exponential kernel.

We aggregate biases as before but replace the edge aggregation function with the following.
\begin{definition}[Local Edge Aggregation Function]\label{def:aggregations_local}
The \emph{local edge  aggregation function} with respect to
the \emph{input} $x$
 is 
\begin{multline*}
    \agg^e_{x}(C_1, C_2) =\\ \sum_{x' \in \dataset'}  \pi_{x',x}
    \sum_{v_{l,i} \in C_1} 
    \frac{1}{|C_2|.\outputf^{\mu}_{x'}
    (v_{C_1})} \sum_{v_{l+1,j} \in C_2} W^l_{i,j}\outputf^\mlp_{x'}(v_{l,i})
    \label{eq:local_explanations_edge_computation}
\end{multline*}
where $\outputf^\mlp_{x'}(v_{l,i})$ is the activation value of the neuron $v_{l,i}$ in the original MLP and $\outputf^\mu_{x'}(v_{
C_1})$ is the activation value of the cluster-neuron $
C_1$ in the clustered MLP. 
\end{definition}
Note that, by this definition, the edge weights are computed layer by layer from input to output.

\begin{example}
\label{ex:localclouds}
Fig.~\ref{fig_XOR_example_summarized}.e shows the clustered MLP for the local explanation of the XOR example (Fig.~\ref{fig_XOR_example_summarized}.a) where $x_0=0$ and $x_1=1$. Fig.~\ref{fig_XOR_example_summarized}.f shows the local explanation as a QAF. The word-cloud representation is also shown in Fig.~\ref{fig_XOR_example_summarized}.g.
In this example, and in general for word-clouds for local explanations,  
we can measure ``relevance'' by the  absolute value of edge weight times activation. 
\end{example}

\section{Desirable 
Properties of Explanations}
To evaluate SpArX, we 
propose three measures 
for assessing
faithfulness and comprehensibility of 
explanations. In this section, we assume as given an MLP \mlp\ of depth $d$ and a corresponding clustered MLP $\mu$.

To begin with, we consider a \emph{faithfulness}
measure inspired by the notion of fidelity considered
for LIME ~\cite{LIME_2016should}, based on 
measuring the \emph{input-output} 
difference  between the original model (in our case, 
$\mlp$) and the 
substitute model (in our case,
the clustered MLP/QAF). 

\begin{definition}[Input-Output Unfaithfulness
]
\label{def:io-faithfulness}
The \emph{local input-output unfaithfulness} of 
$\mu$ to 
$\mlp$ with respect
to \emph{input}  $x$ and \emph{dataset $\dataset$}
is 
  \begin{equation*}
     \mathcal{L}^{\mlp}(\mu) = \sum_{x' \in \dataset} \pi_{x',x} \sum_{v \in V_{d+1}}(\outputf_{x'}^\mlp(v) - \outputf_{x'}^\mu(v))^2.
    \end{equation*}
The \emph{global input-output unfaithfulness} of $\mu$ to $\mlp$ with respect
to dataset 
\emph{$\dataset$} is 

 \begin{equation*}
     \mathcal{G}^{\mlp}(\mu) = \sum_{x' \in \dataset}  \sum_{v \in V_{d+1}}(\outputf_{x'}^\mlp(v) - \outputf_{x'}^\mu(v))^2.
\end{equation*}    
\end{definition}

The lower the input-output unfaithfulness of the clustered MLP $\mu$, the more  faithful $\mu$ is to the original MLP.  

The input-output unfaithfulness measures deviations in the input-output behaviour of the substitute model, 
but, since clustered MLPs maintain
much of the structure of the original MLPs,
we can define a more fine-grained notion of \emph{structured faithfulness} 
by comparing the outputs of the individual
neurons 
in the MLP with the outputs of the 
cluster-neurons summarizing them in the clustered MLP. 

\begin{definition}[Structural Unfaithfulness]
\label{def:structfaithfulness}
Let $K_l$ be the number of clusters at hidden layer $l$ 
 in $\mu$ ($0\!\!<\!\!l\!\!\leq \!\! d$) and $K_{d+1}$ be the number of output neurons. Let $K_{l, j}$ be the number of neurons in the $j$th cluster-neuron $C_{l,j}$ 
 ($0\!<\!l\!\leq \!d+1$, with $K_{d+1,j}\!=\!1$
 ). The
\emph{local structural unfaithfulness}  
 of $\mu$ to $\mlp$ with respect to \emph{input}  $x$ and \emph{dataset} $\dataset$
is:
    \begin{equation*}
    \mathcal{L}^{\mlp}_s(\mu) = \sum_{x' \in \dataset} \pi_{x',x}\sum_{l=1}^{d+1}\sum_{j=1}^{K_l}\sum_{v_{l,i} \in C_{l,j}}(\outputf_{x'}^\mlp(v_{l,i}) - \outputf_{x'}^\mu(C_{l,j}))^2. 
    \label{eq:structural_fidelity}
\end{equation*}

The \emph{global structural unfaithfulness} $\mathcal{G}^{\mlp}_s(\mu)$
is defined analogously by removing the similarity terms
$\pi_{x',x}$.\footnote{See the SM in \cite{ayoobi2023} for the formal definition
.}
\end{definition}

The lower the structured unfaithfulness of the clustered MLP $\mu$, the more  structurally faithful $\mu$ is to the original MLP.  
Note that our notion of structural faithfulness is different from the notions of structural descriptive accuracy by \cite{SUM22}:  they characterise bespoke explanations, defined therein, of probabilistic classifiers equipped with graphical structures and cannot be used in place of our notion, tailored to local and global explanations with SpArX.

Finally, we consider the \emph{cognitive complexity} of explanations based on their
size, inspired by
the  cognitive tractability notion in \cite{Cyras_19}. We use the number of cluster-neurons/arguments 
as a measure.
\begin{definition}[Cognitive Complexity]
\label{def:compl}
Let $K_l$ be the number of clusters at hidden layer $l$ 
 in $\mu$ ($0<l\leq d$). 
 Then, the \emph{cognitive complexity} of $\mu$
 is defined as
 \begin{equation*}
  \Omega(\mu) = \underset{0<l<d+1}{\Pi}K_l.
     \label{eq:cognitive_complexity_general}
 \end{equation*}
\end{definition}

Note that there is a tradeoff between
faithfulness and cognitive complexity
. By reducing the number of cluster-neurons, we reduce cognitive complexity.
However, this also results in higher variance in the neurons 
summarized in the clusters, so the faithfulness of the explanation may suffer. We will explore this trade-off 
in Section~\ref{sec:cognitive}.

Finally,
note that
other properties of explanations by symbolic approaches, notably by \cite{Amgoud22},
are unsuitable for our mechanistic explanations as QAFs. 
Indeed, these existing properties 
focus on
the input-output behaviour of classifiers, rather than their mechanics
.




\section{Experiments}


We conducted four sets of experiments to 
evaluate SpArX with respect to (i) the trade-off between its
sparsification and its ability
to maintain 
faithfulness (Section~\ref{sec:globaleval} for global and Section~\ref{sec:localeval} for local explanations),
and (ii) SpArX's scalability (Section~\ref{sec:scalability})
.
%
We 
used four datasets for classification: 
{\tt iris}
\footnote{\href{https://archive.ics.uci.edu/ml/datasets/iris}{https://archive.ics.uci.edu/ml/datasets/iris}} with 150 instances, 4 continuous features 
and 3 classes;  {\tt breast cancer}
\footnote{\href{https://archive.ics.uci.edu/ml/datasets/breast+cancer+wisconsin+(diagnostic)}{https://archive.ics.uci.edu/ml/datasets/cancer}} with 569 instances, 30 continuous features and 2 classes; {\tt COMPAS}
~\cite{COMPAS} with 11,000 instances and 52 categorical features, to classify \textit{$two\_year\_recid$}; and
{\tt forest covertype}
\footnote{\href{https://archive.ics.uci.edu/ml/datasets/covertype}{https://archive.ics.uci.edu/ml/datasets/covertype}}~\cite{BLACKARD1999131} with 581,012 instances, 54 features (10 continuous, 44 binary
), and 7 classes. 

The first three datasets are standard benchmarks in the 
literature, from three
different domains (biology/medicine/law). 
We used these datasets to evaluate the (input-output and structural) unfaithfulness of the global and local explanations generated by SpArX. 
These datasets however only require small MLPs (see the SM in \cite{ayoobi2023}).
We then used the last
dataset, which is another standard benchmark but 
requires 
deeper MLPs 
with more hidden neurons (see the SM in \cite{ayoobi2023}), to evaluate the scalability of SpArX
. 

For the experiments with the first three datasets, we used  MLPs with 2 hidden layers and 50 hidden neurons each, whereas for the experiments with the last dataset 
we used 1-5 hidden layers with 100, 200 or 500  neurons
. For all experiments, we used the RELU activation function for the hidden neurons and softmax for the output neurons. We give classification performances for all MLPs and average run-times 
for generating local explanations in the SM in \cite{ayoobi2023}.

When experimenting with SpArX, one needs to choose the number of clusters/cluster-neurons at each hidden layer: we do so by specifying a \emph{compression ratio} (for example, a compression ratio of 0.5 amounts to obtaining half cluster-neurons than the original neurons).

\subsection{Global Faithfulness (Comparison to HAP) 
}
\label{sec:globaleval}
Since SpArX essentially compresses an MLP to construct a clustered MLP/QAF
,
one may ask how 
it compares to existing compression approaches.\footnote{
Whereas existing NN compression methods typically
retrain after compression,  we 
do not, as we want to explain the original
NN.} To assess the faithfulness of our global explanations, we  compared SpArX's clustering approach to the state-of-the-art compression
method Hessian Aware Pruning (HAP) ~\cite{Yu_2022_WACV}, which uses relative Hessian traces to prune insensitive parameters in NNs. We measured both input-output and structural unfaithfulness of SpArX and HAP to the original MLP, using the result of HAP compression in place of $\mu$ when applying Definitions~\ref{def:io-faithfulness} and \ref{def:structfaithfulness} for comparison.

\paragraph{Input-Output Faithfulness. 
}   
Table \ref{tab:fidelity} shows the input-output unfaithfulness of global explanations (
$\mathcal{G}^{\mlp}(\mu)$ in Definition~\ref{def:io-faithfulness}) obtained from SpArX 
and HAP using the three chosen datasets and different compression ratios. The unfaithfulness of global explanations in SpArX is lower than HAP, especially for high compression ratios. Note that this does not mean that SpArX is a better compression method, but that the compression method in SpArX is better for 
our purposes (i.e., compressing the MLP while
keeping its mechanics).

\paragraph{Structural Faithfulness.}
%
Table \ref{tab:structural_fidelity} gives the structural global unfaithfulness ($\mathcal{G}^{\mlp}_s(\mu)$ in Definition~\ref{def:structfaithfulness}) for SpArX and HAP, on the three chosen datasets, using different compression ratios
. 
Our method has a much lower structural 
unfaithfulness than HAP 
by preserving activation values close to the original model.

\begin{table}
    \centering
\begin{tabular}{ccccc} \toprule
      \multirow{2}{*}{Method}   & {Compression} & \multicolumn{3}{c}{Datasets} \\ \cmidrule{3-5}
     & {Ratio} & {Iris} & {Cancer} & {COMPAS} \\ \midrule
    HAP  & \multirow{2}{*}{0.2} & 0.05 & 0.48 & 0.02  \\
     SpArX 
     &   & \textbf{0.00} & \textbf{0.02}  & \textbf{0.00}  
       \\ \midrule
     HAP & \multirow{2}{*}{0.4}  & 0.23 & 0.53 & 0.11  \\
     SpArX 
     &   & \textbf{0.00} & \textbf{0.05}  & \textbf{0.00}  
       \\ \midrule
     HAP   & \multirow{2}{*}{0.6}  & 0.23 & 0.58   & 0.20  \\
     SpArX 
     &   & \textbf{0.00} & \textbf{0.10}  & \textbf{0.00}       \\ \midrule
     HAP  & \multirow{2}{*}{0.8}  & 0.28 & 1.00  & 0.26  \\
    SpArX 
    &    & \textbf{0.00} & \textbf{0.21}   & \textbf{0.00}   \\ \bottomrule
\end{tabular}
    \caption{Global input-output unfaithfulness of sparse
    MLPs generated by HAP vs our SpArX method. (Best results in {\bf bold})}
    \label{tab:fidelity}
\end{table}
\begin{table}
    \centering

\begin{tabular}{ccccc} \toprule
      \multirow{2}{*}{Method}   & {Compression} & \multicolumn{3}{c}{Datasets} \\ \cmidrule{3-5}
     & {Ratio} & {Iris} & {Cancer} & {COMPAS} \\ \midrule
    HAP  & \multirow{2}{*}{0.2} & 0.23 & 9.54 & 0.10  \\
     SpArX 
     &   & \textbf{0.00} & \textbf{0.83}  & \textbf{0.02}  
       \\ \midrule
     HAP & \multirow{2}{*}{0.4}  & 0.89 & 61.57 & 0.24  \\
     SpArX 
     &   & \textbf{0.00} & \textbf{1.04}  & \textbf{0.03}  
       \\ \midrule
     HAP   & \multirow{2}{*}{0.6}  & 1.37 & 61.57   & 0.46  \\
     SpArX 
     &   & \textbf{0.00} & \textbf{1.40}  & \textbf{0.04}       \\ \midrule
     HAP  & \multirow{2}{*}{0.8}  & 3.00 & 116.20 & 1.20  \\
    SpArX 
    &    & \textbf{0.02} & \textbf{2.34}   & \textbf{0.05}   \\ \bottomrule
\end{tabular}
    \caption{Global structural unfaithfulness of sparse
    MLPs generated by HAP vs our  SpArX method.}
    \label{tab:structural_fidelity}
\end{table}



\subsection{
Local 
Faithfulness (Comparison to LIME) 
}
\label{sec:localeval}
In order to evaluate 
the local input-output unfaithfulness of SpArX ($\mathcal{L}^{\mlp}(\mu)$ in Definition~\ref{def:io-faithfulness}), we compared SpArX
 and LIME ~\cite{LIME_2016should}\footnote{\url{https://github.com/marcotcr/lime}}, which approximates a target point locally by
 an interpretable substitute model.\footnote{We used ridge regression,  suitable with tabular data in LIME. We used the substitute model as $\mu$ when applying Definition~\ref{def:io-faithfulness} to LIME.}
 Table \ref{tab:local_fidelity} shows the input-output unfaithfulness of the local explanations for LIME and SpArX. We used the same sampling approach as LIME ~\cite{LIME_2016should}. We averaged the unfaithfulness measure for all test examples. The results show that the local explanations produced by our approach are more input-output faithful to the original model. Thus, basing local explanations on 
keeping the MLP mechanics helps also with their input-output faithfulness. 

\begin{table}
    \centering
\begin{tabular}{cccc} \toprule
      \multirow{2}{*}{Method}   & \multicolumn{3}{c}{Datasets} \\ \cmidrule{2-4}
      & {Iris} & {Cancer} & {COMPAS} \\ \midrule
    LIME   & 0.3212 & 0.1623 & 0.0224  \\
     SpArX (0.6)   & \textbf{0.0257} & \textbf{0.0055}  & \textbf{0.0071}  
       \\
       SpArX (0.8)   & \textbf{0.0707} & \textbf{0.0156}  & \textbf{0.0083} \\\bottomrule
\end{tabular}
    \caption{Local input-output unfaithfulness of LIME vs our SpArX method (with different compression ratios, in brackets).}
    
    \label{tab:local_fidelity}
\end{table}


\begin{table}
    \centering
\begin{tabular}{ccccc} \toprule
      \multirow{2}{*}{$\#$Layers}  &       \multirow{2}{*}{Method}  & 
      \multicolumn{3}{c}{$\#$Neurons} \\ \cmidrule{3-5} &
      & {100} & {200} & {500} \\ \midrule
    1 & LIME    & 0.2375 & 0.2919 & 0.3123  \\
     1 & SpArX    & \textbf{0.0000} & \textbf{0.0018}  & \textbf{0.0000}  
       \\\midrule
       2 & LIME   & 0.2509 & 0.2961  & 0.3638
       \\
       2 & SpArX    & \textbf{0.0019} & \textbf{0.0015}  & \textbf{0.0034}
       \\\midrule
       3 & LIME    & 0.3130 & 0.3285  & 0.3127
              \\
       3 & SpArX    & \textbf{0.0028} & \textbf{0.0026}  & \textbf{0.0000}
              \\\midrule
       4 & LIME    & 0.3395 & 0.3459  & 0.3243 
       \\
       4 & SpArX    & \textbf{0.0001} & \textbf{0.0049}  & \textbf{0.0000}
       \\\midrule
       5 & LIME    & 0.3665 & 0.3178  & 0.3288
              \\
       5 & SpArX   & \textbf{0.0030} & \textbf{0.0064}  & \textbf{0.0000}
       \\\bottomrule
\end{tabular}
    \caption{Evaluating scalability of SpArX ({\tt forest covertype} dataset): local input-output unfaithfulness of SpArX (with $80\%$ compression ratio) and LIME using various MLPs 
    with different numbers of hidden layers ($\#$Layers) and 
    neurons ($\#$Neurons)
    .}
    
    \label{tab:MLP_covertypes_80}
\end{table}

\subsection{Scalability}\label{sec:scalability}

To evaluate the scalability of SpArX, we measured its input-output faithfulness on MLPs of increasing complexity, in comparison with 
 LIME ~\cite{LIME_2016should}, using  {\tt forest covertype}  
as a sufficiently large dataset to be tested with various MLP architectures of different sizes
. We have trained 15 MLPs with varying numbers of hidden layers ($\#$Layers) and different numbers of hidden neurons ($\#$Neurons) at each hidden layer (see details in the SM in \cite{ayoobi2023}).

Table \ref{tab:MLP_covertypes_80} 
compares the input-output unfaithfulness of the local explanations by SpArX using $80\%$ compression ratio\footnote{For 
experiments with lower compression ratios, 
see the SM in \cite{ayoobi2023}.} with LIME, all averaged over the test set. The 
results confirm that SpArX explanations are scalable to different MLP architectures of different sizes
.

\section{Towards Tailoring SpArX to Users
}
\label{sec:cognitive}

\begin{figure}[t]
\centering
  \begin{subfigure}{.35\linewidth}
  \centering
  \includegraphics[height=8cm]{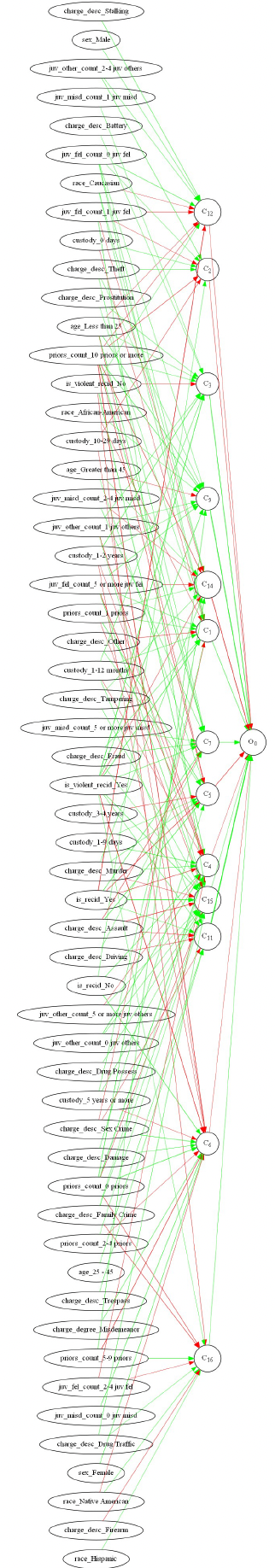}
  \caption{$20\%$ compression ratio}
  \label{fig_compas_global:sub0}
\end{subfigure}%
\begin{subfigure}{.70\linewidth}
  \centering
  \hspace{-1.3cm}
  \includegraphics[width=1.2\linewidth]{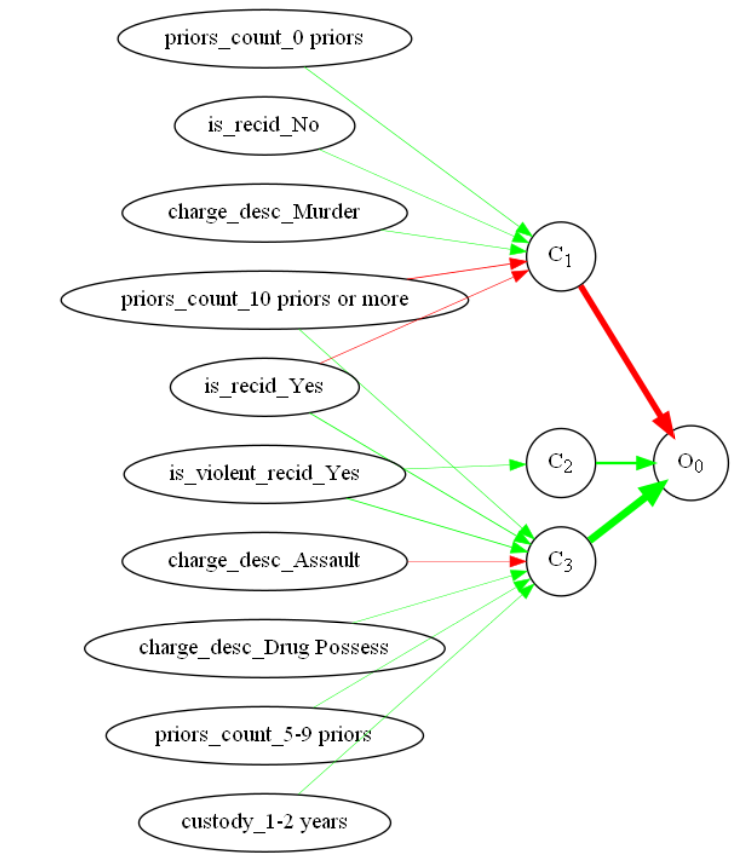}
  \caption{$85\%$ compression ratio}
  \label{fig_compas_global:sub1}
\end{subfigure}%
\caption{Global explanations by SpArX of an MLP with $20\%$ and $85\%$ compression ratios for  {\tt COMPAS}. Here $O_0$ concerns recommitting a crime after two years.
(Sub-figure (a) is given to emphasize poor interpretability due to size, so readability is not a concern).} 
\label{fig_compas_global}
\end{figure}

\begin{figure}[t]
  \begin{subfigure}{.35\linewidth}
  \centering
  \hspace{-4mm}
  \includegraphics[width=1.1\linewidth]{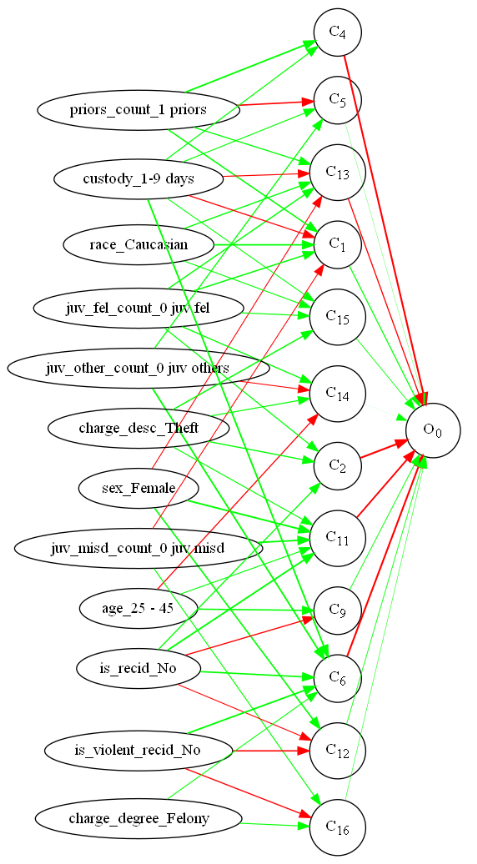}
  \caption{$20\%$ compression ratio}
  \label{fig_compas_local:sub0}
\end{subfigure}%
\begin{subfigure}{.7\linewidth}
  \centering
  \includegraphics[width=1.0\linewidth]{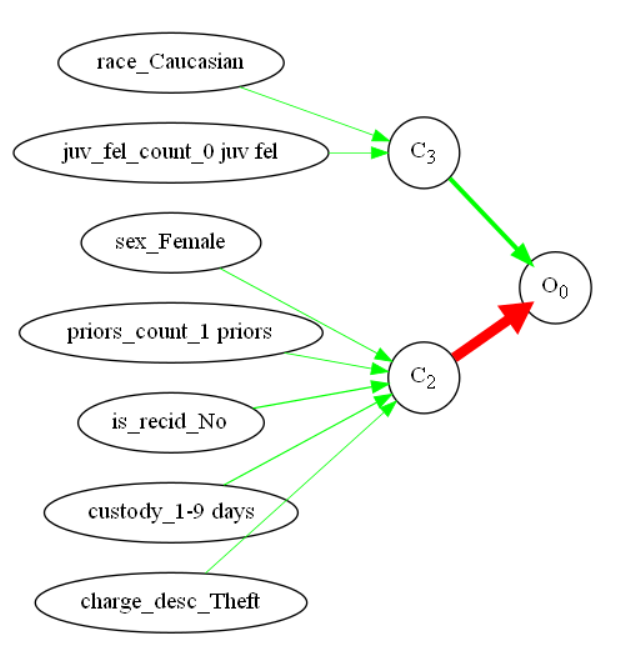}
  \caption{$85\%$ compression ratio}
  \label{fig_compas_local:sub1}
\end{subfigure}%
\caption{Local explanations by SpArX of an MLP with $20\%$ and $85\%$ compression ratios for {\tt COMPAS}. Here $O_0=0$ suggests that the individual 
will not be recommitting a crime after two years. (
Again, the readability of sub-figure (a) is not a concern).}
\label{fig_compas_local}
\end{figure}

The explanations obtained with SpArX are in the form of QAFs drawn from the sparsified MLPs, in the spirit of much work in argumentative explainable AI~\cite{argXAIsurvey}. 
In this section, we explore how  they might be tailored to 
users. We first consider the property of cognitive complexity for SpArX 
(see Definition~\ref{def:compl})
and the tradeoff between faithfulness and cognitive complexity (Section~\ref{sec:actual cognitive})
and then illustrate how local and global explanations can be the starting point to obtain more natural explanations
(Section~\ref{sec:human}).  
Throughout this section, we focus on examples only, all drawn in the context of an MLP trained on the {\tt COMPAS} dataset with one hidden layer and 20 
neurons in the hidden layer (see details in the SM in \cite{ayoobi2023}). 
We chose {\tt COMPAS} because it is a very popular dataset in the literature, and it is the largest, amongst those we consider,  for binary classification.

\subsection{Cognitive Complexity}
\label{sec:actual cognitive}
The cognitive complexity of SpArX 
depends on the number of clusters per layer. Fewer clusters lead to a more interpretable explanation at the cost of achieving lower (structural) faithfulness.

Fig.~\ref{fig_compas_global:sub0} shows a {\em global explanation} for the given MLP for  {\tt COMPAS}, 
with $20\%$ compression ratio and pruning edges  with low weights.\footnote{ Note that pruning is only done here for visualization.} 
The classification results  of the clustered MLP underpinning this explanation is $98.32\%$, the same as the original MLP.
This explanation is clearly hard to interpret by a user.    
%
Fig.~\ref{fig_compas_global:sub1} shows the global explanation 
with $85\%$ compression rate and, again, pruning the edges with low weights
. This global explanation is more comprehensible 
(and 
the classification results are $94.60\%$, the same as the original model). 

Using the same MLP for the {\tt COMPAS} dataset, {\em local explanations} of an input example, with  $20\%$ and $85\%$ compression ratios, are shown in Fig.~\ref{fig_compas_local} (both clustered MLPs compute the same output $O_0=0$, indicating that the input individual, with the features as given in the figure, is predicted to not re-offend within two years). 
Fig.~\ref{fig_compas_local:sub1} is more interpretable than Fig.~\ref{fig_compas_local:sub0}
, but the two 
clustered MLPs make the same prediction for the given input, faithfully to the MLP.

\subsection{From SpArX to Natural Explanations}
\label{sec:human}

Global and local explanations obtained by SpArX can be presented so that humans can progressively inquire about the reasoning of the underlying MLP, e.g. by instantiating templates as in \cite{ CocarascuRT19,Cyras_19}.
%
%
For illustration, consider the global explanation in Fig.~\ref{fig_compas_global:sub1} again.
Unlike shallow input-output explanations, we can see the role of each hidden cluster-neuron in the proposed method. 
There are two sets of hidden cluster-neurons, namely an attacker ($C_1$) and two supporters ($C_2$ and $C_3$). 
They could be shown incrementally, following prompts from a human user, to explain the output. 
Specifically,
$C_1$ attacks the output, indicating that the input individual will not recommit a crime in a two-year period. Three features are supporting $C_1$ and two features are attacking it, and they could also be shown to a human user on demand. The attacking features also support $C_3$. This means that they strengthen the support by $C_3$ and weaken the attack by $C_1$. Therefore, \textit{priors\_count\_10 priors or more} and \textit{is\_recid\_Yes} both strongly affect the output. Indeed, looking at the {\tt COMPAS} dataset, more than $99\%$ of individuals that have these two features recommitted the crime in a two years period. $C_2$ and $C_3$ are supporting the output. $C2$ is only supported by the \textit{is\_violent\_recid\_Yes} feature. This suggests that if individuals have a violent recidivism they are likely to recommit a crime after two years. In the {\tt COMPAS} dataset, this conjecture is $100\%$ valid. These kinds of interpretations go beyond the shallow input-output explanations offered 
e.g. by LIME, but can be automatically drawn from the QAFs generated by SpArX. $C3$ is supported by several features and is attacked by one feature. Looking at this argumentative global explanation one can understand the effect of each feature 
as well as of each hidden neuron on the output.

Similar considerations can be made for local explanations, e.g. 
the explanation in Fig.~\ref{fig_compas_local:sub1}.
The class label for this input example is 0, which means that the individual has not recommitted the crime in a two-year period. The local explanation shows this fact by emphasising $C_2$ as a stronger attacker than the supporter $C_3$. $C_2$ says that since the sex of the individual is female, she has only 1 prior, she has no recidivism, she has custody time of 1 to 9 days and the crime was theft, she will not recommit the crime in a two years period. 

\begin{figure}[H]
\centering
\begin{subfigure}{1\linewidth}
  \centering
  \includegraphics[width=1.0\linewidth]{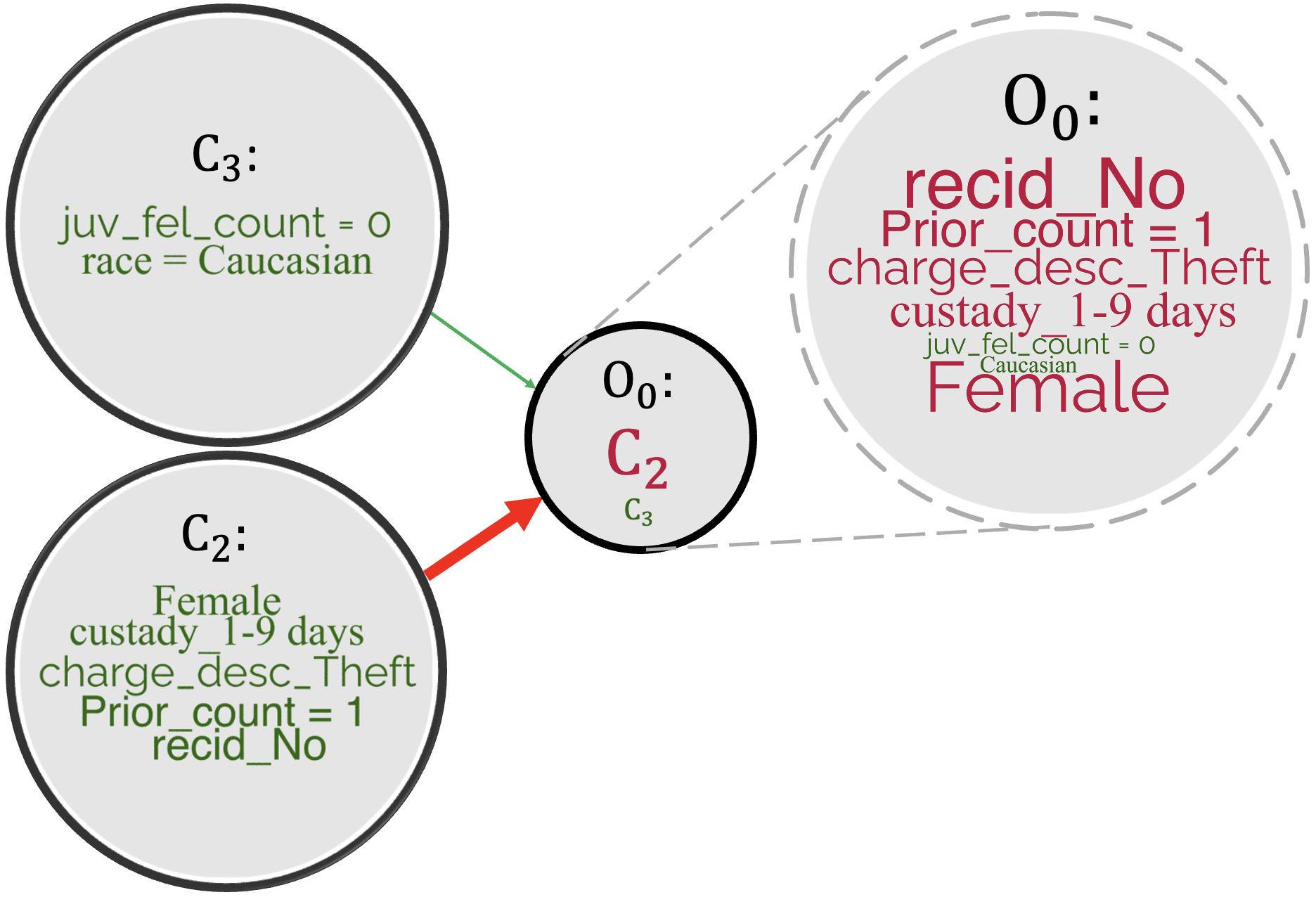}
  \label{fig_wc_fig3}
\end{subfigure}%
\caption{Word-cloud presentation of the cluster-neurons and output neuron for the local explanation in Fig.~\ref{fig_compas_local:sub1}
. To enhance comprehensibility, we can 
display the word-cloud of the output node with respect to the input features instead of the 
clusters 
(see dashed lines).}  
\label{fig_wc_compas}
\end{figure}

For these readings of explanations to be natural,    human-interpretable presentations of the cluster-neurons are needed. Specifically, we could use the word-cloud presentation from Section~\ref{sec:sparx} (see Examples~\ref{ex:globalclouds}  and \ref{ex:localclouds}).
For illustration, 
Fig.~\ref{fig_wc_compas} shows the local explanation 
in Fig.~\ref{fig_compas_local:sub1}
with word-clouds for presenting hidden cluster-neurons and output neurons.\footnote{See the SM in \cite{ayoobi2023} for the  word-cloud variant of the global explanation in Fig.~\ref{fig_compas_global:sub1}.} 
Other ways to present cluster-neurons may be useful: we leave the exploration of alternatives as well as human studies to assess their amenability to humans
to future work.

\section{Conclusion} 
We introduced SpArX, a novel method for generating
argumentative explanations for MLPs.
In contrast to shallow input-output explainers like LIME, SpArX  maintains
structural similarity to the original MLP
in order to give faithful explanations, while allowing tailoring them to 
comprehensibility for users.
Our experimental results show that the explanations 
by
SpArX are more 
faithful to the original model than LIME. We have also compared SpArX with a state-of-the-art NN compression technique called HAP
, showing
that SpArX is 
more faithful to the original model
.
 Further, our 
 explanations are more \emph{structurally} faithful to the original model by providing 
 deeper insights into the mechanics thereof
 , and can be tailored to users for cognitive tractability and to obtain natural explanations. 

Future research includes extending 
SpArX to other types of NNs, e.g. CNNs, as well as furthering it to cluster neurons across hidden layers. 
It would also be interesting to explore whether SpArX could be extended to exploit formal relationships between NNs and other symbolic approaches, 
e.g. in~\cite{Giordano2021}.
Further, it would be interesting to explore formalizations such as in~\cite{Lorini21} for characterizing  
uncertainty as captured by SpArX
.


\section*{Acknowledgments}
This research was partially funded by the  European Research Council (ERC) under the
European Union’s Horizon 2020 research and innovation programme (grant
agreement No. 101020934, ADIX) and by J.P. Morgan and by the Royal
Academy of Engineering under the Research Chairs and Senior Research
Fellowships scheme.  Any views or opinions expressed herein are solely those of the authors.

\bibliographystyle{named}
\bibliography{main}

\end{document}


\begin{frontmatter}

\title{Supplementary Materials: \\SpArX: Sparse Argumentative Explanations \\ for Neural Networks}

\author[A]{\fnms{Hamed}~\snm{Ayoobi}\orcid{0000-0002-5418-6352}\thanks{Corresponding Author. Email: h.ayoobi@imperial.ac.uk}}
\author[B,A]{\fnms{Nico}~\snm{Potyka}}
\author[A]{\fnms{Francesca}~\snm{Toni}}

\address[A]{Department of Computing, Imperial College London, United Kingdom}
\address[B]{Cardiff University, United Kingdom}
\end{frontmatter}


\section{Foundations}

\subsection{Proof of Proposition \ref{prop:translation}}

\setcounter{proposition}{0}
\begin{proposition}
\label{prop:translation}
Let $\mlp$ be an MLP with activation function $\varphi$ 
that is invertible or ReLU. Then,  for  every input $x$, 
the 
QAF $Q_{\mlp,x}$ satisfies 
$\outputf^\mlp_{x}(v_{l,i}) = \strength(A_{l,i})$, where
$\strength(A_{l,i})$ denotes the final
strength of $A_{l,i}$ in $Q_{\mlp,x}$.
\end{proposition}
\begin{proof}
$Q_{\mlp,x}$ is acyclic by construction. 
For acyclic QAFs, the iterative two-step update procedure is equivalent to a simple forward pass, where 
each argument is updated once according to any topological 
ordering of the arguments
~\cite{potyka_modular_2019}.
Given that $Q_{\mlp,x}$ 
has the graphical structure of $\mlp$, 
a natural topological
ordering is given by $A_{l,i} < A_{l',i'}$ if 
$l\leq l'$ and $i < i'$. The update order then corresponds to
the standard forward propagation procedure in MLPs.
Then, the claim follows by induction over the depth $d$ of $\mlp$. 
For the input arguments at depth $d=0$, we do not have 
any attackers or supporters. Therefore, the aggregation function will
just return $0$ and we have 
$\strength(A_{0,i}) =
\varphi\big(\varphi^{-1}(x_i) \big) = x_i 
= \outputf^\mlp_{x}(v_{0,i}),
$
which proves the base case.
For the induction step, consider an argument at depth $d+1$. We have
$\strength(A_{d+1,i}) =
\varphi\big(\varphi^{-1}(\varphi(b^{d+1}_i)) + 
\sum_{(A_{d,i'}, A_{d+1,i}) \in E} W^d_{i',i} \cdot \strength(A_{d,i'}) \big)
= 
\varphi\big(b^{d+1}_i + 
\sum_{(A_{d,i'}, A_{d+1,i}) \in E} W^d_{i',i} \cdot \outputf^\mlp_{x}(v_{d,i'}) \big)
= \outputf^\mlp_{x}(v_{d+1,i}),
$ which completes the proof.
\end{proof}

\subsection{Motivation of Aggregation Functions}

Our global aggregation function is motivated as 
follows. Every neuron in the original MLP $\mlp$
is represented by a cluster in the corresponding 
clustered MLP $\mu$.
Ideally, we want that the output of the cluster 
in $\mu$ is equal to the output of all neurons
in $\mlp$. However, obviously, this is only
possible if all neurons in the cluster would have 
the same output in the first place.
The next best
thing to do is then to minimize the expected deviation.
However, due to the layerwise dependencies
(the activation of a neuron depends on all weights and biases in all previous layers) and non-linear activation functions,
the resulting objective function does not possess
a simple closed-form solution and the solution is
not necessarily unique. We, therefore, choose a heuristic
approach to define the aggregation functions that 
should result in a small, but not necessarily minimal deviation. 

Instead of trying to minimize the deviation in the activations of cluster neurons and neurons in the cluster directly, we minimize the deviation of the
parameters (weights and biases) that cause the deviation in the activations. For the particular case that all neurons in 
a cluster have equal parameters (and therefore equal outputs),
this minimization would result in the common parameters 
and the deviation would be $0$. Based on these considerations, we now derive the
global bias and weight aggregation functions.

For biases, given a cluster neuron $v_C$
with corresponding bias $b$, we want to minimize
\begin{equation}
\sum_{v_{l,i} \in C} (b - b^l_i)^2.    
\end{equation}
As usual, we use the fact that the partial derivatives at a minimum must be $0$.
By setting the derivative with respect to $b$
to $0$ and solving for $b$, we find that we must
have
\begin{equation}
    b = \frac{\sum_{v_{l,i} \in C} b^l_i}{|C|},
\end{equation}
which corresponds to our bias aggregation function.

Thinking about the weights is slightly more complicated
because the weights represent the connection strength
between two cluster neurons. When considering an edge
$(v_{C_1}, v_{C_2})$ in the clustered MLP, we have to take account of the fact that every neuron 
$v_2 \in C_2$ is potentially influenced by all neurons
in the previous layer. In particular, it is potentially
influenced by all $v_1 \in C_1$. Hence, while $v_2$
will potentially receive $|C_1|$ inputs from the 
neurons in $C_1$, $v_{C_2}$ will receive only one
input from $v_{C_1}$. When comparing the weight
$w$ of the edge $(v_{C_1}, v_{C_2})$ with a
weight $W^l_{j,i}$ between neurons $v_{l,i} \in C_1$
and $v_{l,j} \in C_2$, it is therefore important to divide
$w$ by $|C_1|$  (since $(v_{C_1}, v_{C_2})$ represents $|C_1|$ edges,
it should be $|C_1|$ times larger than the 
weight of the individual edges).
Hence, given a pair of clusters
$(C_1, C_2)$ with corresponding weight $w$, we want to minimize
\begin{equation}
 \sum_{v_{l,i} \in C_1}  
 \sum_{v_{l+1,j} \in C_2}
 (W^l_{j,i} - \frac{w}{|C_1|})^2.
\end{equation}
Setting the derivative with respect to $w$ to $0$
yields the necessary condition
\begin{equation}
    0 = -2 \cdot \sum_{v_{l,i} \in C_1}  
 \sum_{v_{l+1,j} \in C_2}
 (W^l_{j,i} - \frac{w}{|C_1|}).
\end{equation}
Dividing both sides of the equation by $2$ and rearranging terms yields
\begin{equation}
    \sum_{v_{l,i} \in C_1}  
 \sum_{v_{l+1,j} \in C_2}
 W^l_{j,i} 
 = \sum_{v_{l,i} \in C_1}  
 \sum_{v_{l+1,j} \in C_2} \frac{w}{|C_1|} \\
 = |C_2| \cdot w.
\end{equation}
Dividing by $|C_2|$ leads to the expression in our weight aggregation function
\begin{equation}
 w =   \frac{1}{|C_2|} \sum_{v_{l,i} \in C_1}  
 \sum_{v_{l+1,j} \in C_2}
 W^l_{j,i} .
\end{equation}

\subsection{Definition of Global Structural Unfaithfulness}
Let $K_l$ be the number of clusters at hidden layer $l$ 
 in $\mu$ ($0\!\!<\!\!l\!\!\leq \!\! d$) and $K_{d+1}$ be the number of output neurons. Let $K_{l, j}$ be the number of neurons in the $j$th cluster-neuron $C_{l,j}$ 
 ($0\!<\!l\!\leq \!d+1$, with $K_{d+1,j}\!=\!1$
 ). The
\emph{global structural unfaithfulness}  
 of $\mu$ to $\mlp$ with respect to \emph{input}  $x$ and \emph{dataset} $\dataset$
is:
    \begin{equation*}
    \mathcal{G}^{\mlp}_s(\mu) = \sum_{x' \in \dataset}\sum_{l=1}^{d+1}\sum_{j=1}^{K_l}\sum_{v_{l,i} \in C_{l,j}}(\outputf_{x'}^\mlp(v_{l,i}) - \outputf_{x'}^\mu(C_{l,j}))^2. 
    \label{eq:structural_fidelity}
\end{equation*}

\section{Experiments}
This section includes more detailed experimental results for 
Section 7. The code to run the experiments is 
part of this submission and will be made publicly available upon acceptance. 

\subsection{Performance of Trained Models}
Tables \ref{tab:MLP_precisions_iris}, 
\ref{tab:MLP_precisions_cancer}, \ref{tab:MLP_precisions_compas}, and \ref{tab:MLP_precisions_forest} 
show the average F1-scores of the trained MLPs using $90\%$ of the data 
for training the models.  Only for 
the {\tt forest covertype} dataset, increasing the number of hidden layers and the number of hidden neurons in an MLP leads to a higher F1-score.  

\begin{table}
    \centering
\begin{tabular}{cccc} \toprule
      \multirow{2}{*}{$\#$Layers}   & 
      \multicolumn{3}{c}{$\#$Neurons} \\ \cmidrule{2-4} 
      & {100} & {200} & {500} \\ \midrule
    1     & 0.9667 & 0.9667 & 0.9667  

       \\
       2    & 0.9667 & 0.9667  & 0.9667
       \\
       3    & 0.9667 & 0.9667  & 0.9667
              \\

       4     & 0.9667 & 0.9667 & 0.9667 
       \\

       5     & 0.9667 & 0.9667  & 0.9667
       \\\bottomrule
\end{tabular}
    \caption{Average F1-scores for trained MLPs with different numbers of hidden layers ($\#$Layers) and hidden neurons ($\#$Neurons) using the {\tt iris} dataset.}
    
    \label{tab:MLP_precisions_iris}
\end{table}

\begin{table}
    \centering
\begin{tabular}{cccc} \toprule
      \multirow{2}{*}{$\#$Layers}   & 
      \multicolumn{3}{c}{$\#$Neurons} \\ \cmidrule{2-4} 
      & {100} & {200} & {500} \\ \midrule
    1     & 0.9349 & 0.9177 & 0.9183  

       \\
       2    & 0.9445 & 0.9349  & 0.8914
       \\
       3    & 0.9082 & 0.9260  & 0.8758
              \\

       4     & 0.8834 & 0.8820 & \textbf{0.9539} 
       \\

       5     & 0.8741 & 0.9013  & 0.9355
       \\\bottomrule
\end{tabular}
    \caption{Average F1-scores for trained MLPs with different numbers of hidden layers ($\#$Layers) and hidden neurons ($\#$Neurons) using the {\tt cancer} dataset.}
    
    \label{tab:MLP_precisions_cancer}
\end{table}

\begin{table}
    \centering
\begin{tabular}{cccc} \toprule
      \multirow{2}{*}{$\#$Layers}   & 
      \multicolumn{3}{c}{$\#$Neurons} \\ \cmidrule{2-4} 
      & {100} & {200} & {500} \\ \midrule
    1     & 0.7009 & \textbf{0.7064} & 0.6996  

       \\
       2    & 0.6934 & 0.6660  & 0.6622
       \\
       3    & 0.6699 & 0.6738  & 0.6690
              \\

       4     & 0.6715 & 0.6561 & 0.6701 
       \\

       5     & 0.6623 & 0.6873  & 0.6714
       \\\bottomrule
\end{tabular}
    \caption{Average F1-scores for trained MLPs with different numbers of hidden layers ($\#$Layers) and hidden neurons ($\#$Neurons) using the {\tt COMPAS} dataset.}
    
    \label{tab:MLP_precisions_compas}
\end{table}

\begin{table}
    \centering
\begin{tabular}{cccc} \toprule
      \multirow{2}{*}{$\#$Layers}   & 
      \multicolumn{3}{c}{$\#$Neurons} \\ \cmidrule{2-4} 
      & {100} & {200} & {500} \\ \midrule
    1     & 0.8035 & 0.8324 & 0.8795  

       \\
       2    & 0.8622 & 0.9068  & 0.9457
       \\
       3    & 0.8862 & 0.9272  & 0.9519
              \\

       4     & 0.9025 & 0.9356 & 0.9507 
       \\

       5     & 0.9100 & 0.9400  & \textbf{0.9529}
       \\\bottomrule
\end{tabular}
    \caption{Average F1-scores for trained MLPs with different numbers of hidden layers ($\#$Layers) and hidden neurons ($\#$Neurons) using the {\tt forest covertype} dataset.}
    
    \label{tab:MLP_precisions_forest}
\end{table}



    

\subsection{Scalability with Other Compression Ratios}
Tables \ref{tab:MLP_architectures_20}, \ref{tab:MLP_architectures_40}, and \ref{tab:MLP_architectures_60} show the unfaithfulness comparison of the local explanations of SpArX with LIME using $20\%$, $40\%$, and $60\%$ compression ratios for the forest covertype dataset, respectively. 


\begin{table}
    \centering
\begin{tabular}{ccccc} \toprule
      \multirow{2}{*}{$\#$Layers}  &       \multirow{2}{*}{Method}  & 
      \multicolumn{3}{c}{$\#$Neurons} \\ \cmidrule{3-5} &
      & {100} & {200} & {500} \\ \midrule
    1 & LIME    & 0.2375 & 0.2919 & 0.3123  \\
     1 & SpArX    & \textbf{0.0000} & \textbf{0.0004}  & \textbf{0.0000}  
       \\\midrule
       2 & LIME   & 0.2509 & 0.2961  & 0.3638
       \\
       2 & SpArX    & \textbf{0.0018} & \textbf{0.0008}  & \textbf{0.0031}
       \\\midrule
       3 & LIME    & 0.3130 & 0.3285  & 0.3127
              \\
       3 & SpArX    & \textbf{0.0017} & \textbf{0.0016}  & \textbf{0.0000}
              \\\midrule
       4 & LIME    & 0.3395 & 0.3459  & 0.3243 
       \\
       4 & SpArX    & \textbf{0.0001} & \textbf{0.0044}  & \textbf{0.0000}
       \\\midrule
       5 & LIME    & 0.3665 & 0.3178  & 0.3288
              \\
       5 & SpArX   & \textbf{0.0004} & \textbf{0.0048}  & \textbf{0.0000}
       \\\bottomrule
\end{tabular}
    \caption{Evaluating scalability of SpArX: local input-output unfaithfulness of SpArX (with $20\%$ compression ratio) and LIME for trained MLPs with different numbers of hidden layers ($\#$Layers) and neurons ($\#$Neurons) using the {\tt forest covertype} dataset.}
    
    \label{tab:MLP_architectures_20}
\end{table}

\begin{table}
    \centering
\begin{tabular}{ccccc} \toprule
      \multirow{2}{*}{$\#$Layers}  &       \multirow{2}{*}{Method}  & 
      \multicolumn{3}{c}{$\#$Neurons} \\ \cmidrule{3-5} &
      & {100} & {200} & {500} \\ \midrule
    1 & LIME    & 0.2375 & 0.2919 & 0.3123  \\
     1 & SpArX    & \textbf{0.0000} & \textbf{0.0006}  & \textbf{0.0000}  
       \\\midrule
       2 & LIME   & 0.2509 & 0.2961  & 0.3638
       \\
       2 & SpArX    & \textbf{0.0012} & \textbf{0.0026}  & \textbf{0.0031}
       \\\midrule
       3 & LIME    & 0.3130 & 0.3285  & 0.3127
              \\
       3 & SpArX    & \textbf{0.0028} & \textbf{0.0021}  & \textbf{0.0000}
              \\\midrule
       4 & LIME    & 0.3395 & 0.3459  & 0.3243 
       \\
       4 & SpArX    & \textbf{0.0001} & \textbf{0.0044}  & \textbf{0.0000}
       \\\midrule
       5 & LIME    & 0.3665 & 0.3178  & 0.3288
              \\
       5 & SpArX   & \textbf{0.0030} & \textbf{0.0045}  & \textbf{0.0000}
       \\\bottomrule
\end{tabular}
    \caption{Evaluating scalability of SpArX: local input-output unfaithfulness of SpArX (with $40\%$ compression ratio) and LIME for trained MLPs with different numbers of hidden layers ($\#$Layers) and neurons ($\#$Neurons) using the {\tt forest covertype} dataset.}
    
    \label{tab:MLP_architectures_40}
\end{table}

\subsection{Run-time}
Table \ref{tab:run-times} 
shows the average run-time in seconds  of SpArX (using different compression ratios) and LIME for generating local explanations. These results show that on average generating a deep local explanation with SpArX takes longer than generating a shallow (input-output) explanation with LIME. In this experiment, we have used an Intel(R) Core(TM) i7-7700HQ CPU $@$ 2.80GHz  with 16GB of RAM using a windows 10 (education) operating system. SpArX explanations do not use GPU or CPU multiprocessing.

\begin{table}
    \centering
\begin{tabular}{ccccc} \toprule
      \multirow{2}{*}{$\#$Layers}  &       \multirow{2}{*}{Method}  & 
      \multicolumn{3}{c}{$\#$Neurons} \\ \cmidrule{3-5} &
      & {100} & {200} & {500} \\ \midrule
    1 & LIME    & 0.2375 & 0.2919 & 0.3123  \\
     1 & SpArX    & \textbf{0.0000} & \textbf{0.0005}  & \textbf{0.0000}  
       \\\midrule
       2 & LIME   & 0.2509 & 0.2961  & 0.3638
       \\
       2 & SpArX    & \textbf{0.0017} & \textbf{0.0015}  & \textbf{0.0034}
       \\\midrule
       3 & LIME    & 0.3130 & 0.3285  & 0.3127
              \\
       3 & SpArX    & \textbf{0.0038} & \textbf{0.0026}  & \textbf{0.0000}
              \\\midrule
       4 & LIME    & 0.3395 & 0.3459  & 0.3243 
       \\
       4 & SpArX    & \textbf{0.0026} & \textbf{0.0048}  & \textbf{0.0000}
       \\\midrule
       5 & LIME    & 0.3665 & 0.3178  & 0.3288
              \\
       5 & SpArX   & \textbf{0.0028} & \textbf{0.0059}  & \textbf{0.0002}
       \\\bottomrule
\end{tabular}
    \caption{Evaluating scalability of SpArX: local input-output unfaithfulness of SpArX (with $60\%$ compression ratio) and LIME for trained MLPs with different numbers of hidden layers ($\#$Layers) and neurons ($\#$Neurons) using the {\tt forest covertype} dataset.}
    
    \label{tab:MLP_architectures_60}
\end{table}

\begin{table}
    \centering
\begin{tabular}{ccccc} \toprule
      \multirow{2}{*}{$\#$Layers}  &       \multirow{2}{*}{Method}  & 
      \multicolumn{3}{c}{$\#$Neurons} \\ \cmidrule{3-5} &
      & {100} & {200} & {500} \\ \midrule
    1 & LIME    & 0.1612 & 0.1633 & \textbf{0.1645}  \\
     1 & SpArX    & \textbf{0.0747} & \textbf{0.1047}  &  0.3999 
       \\\midrule
       2 & LIME   & \textbf{0.1612} & \textbf{0.1633} & \textbf{0.1645}
       \\
       2 & SpArX    & 0.1865 & 0.2269  & 0.8985
       \\\midrule
       3 & LIME    &\textbf{0.1612} & \textbf{0.1633} & \textbf{0.1645}
              \\
       3 & SpArX    & 0.2881 & 0.3809  & 1.5887
              \\\midrule
       4 & LIME    &\textbf{0.1612} & \textbf{0.1633} & \textbf{0.1645}
       \\
       4 & SpArX    &  0.3430 & 0.4368  & 2.5342
       \\\midrule
       5 & LIME     &\textbf{0.1612} & \textbf{0.1633} & \textbf{0.1645}
              \\
       5 & SpArX   &  0.3729 & 0.5275  & 3.3333
       \\\bottomrule
\end{tabular}
    \caption{Average run-times in seconds (over different compression ratios $20\%$, $40\%$, $60\%$, $80\%$) of SpArX and LIME (in seconds) for generating local explanations from MLPs with different numbers of hidden layers ($\#$Layers) and 
    neurons ($\#$Neurons) for the {\tt forest covertype} dataset.}
    
    \label{tab:run-times}
\end{table}

\section{Cognitive Complexity Experiments}
The F1-score for a trained MLP on the {\tt COMPAS} dataset with a hidden layer and 20 hidden neurons is reported in table \ref{tab:MLP_compas_cognitive}.

\begin{table}
    \centering
\begin{tabular}{ccc} \toprule
      $\#$Layers  & 
      $\#$Neurons & F1-score \\ \midrule 
      1 & 20 & 0.6812        
       \\\bottomrule
\end{tabular}
    \caption{F1-score for a trained MLP with a hidden layer ($\#$Layers) and 20 hidden neurons ($\#$Neurons) using the {\tt COMPAS} dataset.}
    
    \label{tab:MLP_compas_cognitive}
\end{table}

The word-cloud representation of the global explanation in Figure 2 in the paper 
is shown in Figure \ref{fig_wc_compas_global} 
here.

\renewcommand{\thefigure}{5}
\begin{figure}[H]
\centering
  \begin{subfigure}{.9\linewidth}
  \centering
  \includegraphics[width=1\linewidth]{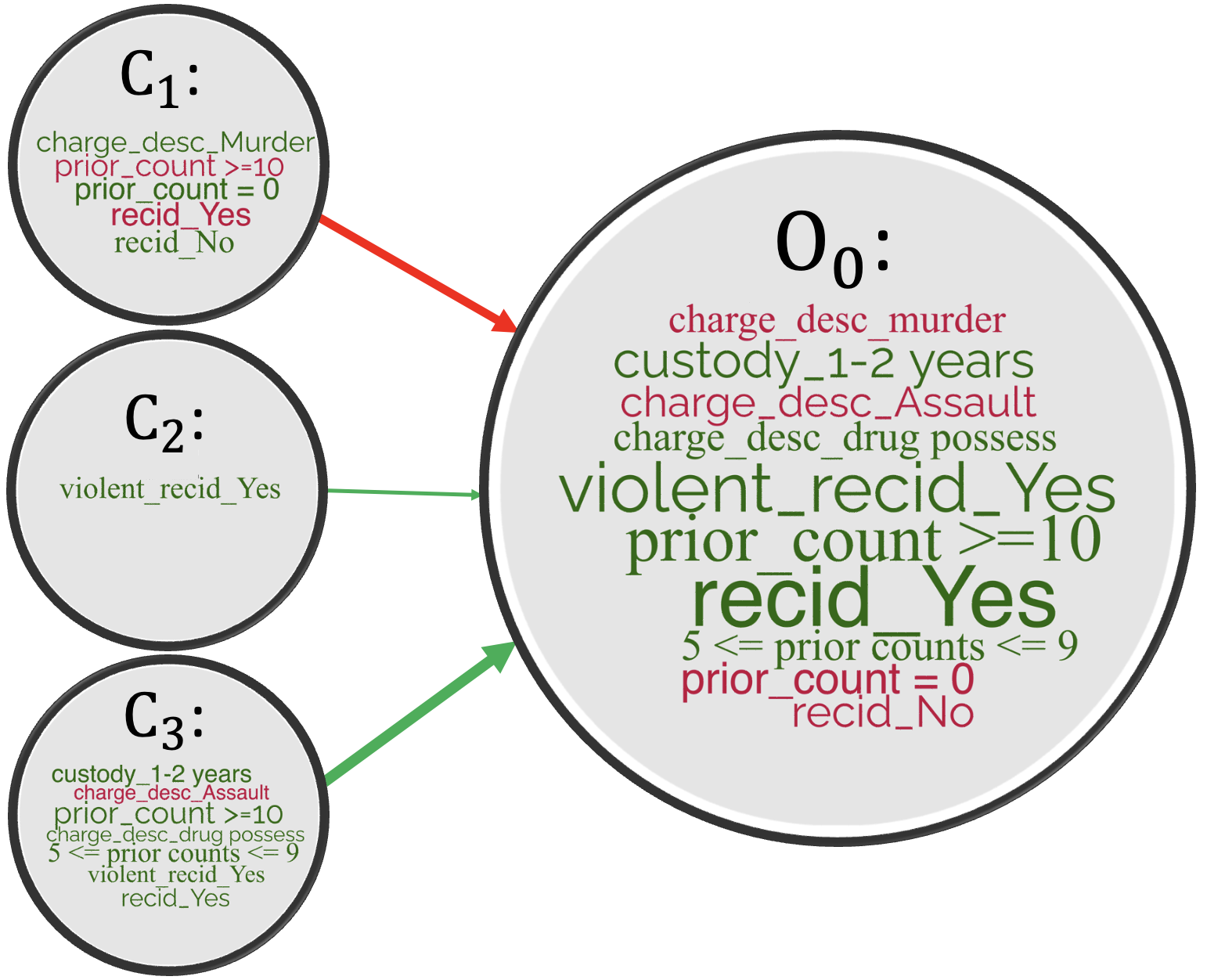}
\end{subfigure}%

\caption{Word-cloud presentation of the cluster-neurons and  output neuron for the global explanation in Figure 2b in the paper 
(presenting the output neuron with respect to the input features, 
for increased interpretability).}
\label{fig_wc_compas_global}
\end{figure}

\bibliographystyle{ecai}
\bibliography{sm}